%% file: main.tex
\documentclass[10pt,twocolumn,letterpaper]{article}

\usepackage{iccv}              % To produce the CAMERA-READY version
\usepackage{dsfont}
\usepackage{xr-hyper}
\usepackage{caption}
\input{preamble}

\def\paperID{5794} % *** Enter the Paper ID here
\def\confName{ICCV}
\def\confYear{2025}

\title{Alias-free 4D Gaussian Splatting\vspace{-0.8em}}

\author{
\small
  Zilong Chen\textsuperscript{1}, % User format: Name*AffilMarker
  Huan-ang Gao\textsuperscript{2},
  Delin Qu\textsuperscript{4}, % User format: Name\textdaggerAffilMarker
  Haohan Chi\textsuperscript{2},
  Hao Tang\textsuperscript{5},
  Kai Zhang\textsuperscript{1,\textdagger}, 
  Hao Zhao\textsuperscript{2,3,\textdagger} \\ 
\tt\small
  \textsuperscript{1}Tsinghua Shenzhen International Graduate School, Tsinghua University \\
\tt\small
  \textsuperscript{2}Institute for AI Industry Research (AIR), Tsinghua University \\
\tt\small
  \textsuperscript{3} Beijing Academy of Artificial Intelligence, BAAI \quad 
    \textsuperscript{4}Fudan University \\ 
\tt\small
    \textsuperscript{5}Peking University \quad
    \textsuperscript{\textdagger}Corresponding author
}

\begin{document}
% \makeatletter
% \let\@oldmaketitle\@maketitle% Store \@maketitle
% \renewcommand{\@maketitle}{\@oldmaketitle% Update \@maketitle to insert...
%   \vspace{-8ex}
%   \begin{center}
%   \captionsetup{type=figure}
%   \setcounter{figure}{0}
%   \includegraphics[trim=0ex 0 0 0, clip, width=1.0\textwidth]{figure/teaser.pdf}
%     \caption{Similar to 3DGS\cite{3dgs}, 4DGS undergoes substantial dilation when the sampling rate is lowered and experiences erosion along with high-frequency artifacts when the sampling rate is increased, as illustrated in (a). The fixed expansion scale in the 3DGS anti-aliasing algorithm Mip-Splatting~\cite{mip-splatting} influences the temporal anisotropy of Gaussians and can render otherwise invisible Gaussians visible, potentially impairing the spatiotemporal perception of 4DGS, as shown in (b). We present \ours, which adjusts the dilation scale based on Gaussian scale variations to mitigate the impact of filtering on 4DGS, while preserving antialiasing capability and maintaining reconstruction quality, effectively eliminating artifacts caused by sampling rate changes.
%     } 
%     %\vspace{1.0em}
%     \label{fig:teaser}
%   \end{center}
% }
%     % 与3DGS\类似，当降低采样率时，4DGS会产生巨大的膨胀，而当增大采样率时，又会产生侵蚀和高频伪影，如(a)中所示，3DGS的抗锯齿算法Mip-Splatting\cite{mip-splatting}固定的膨胀尺度会影响高斯随时间变化过程中的各向异性，同时让不可见的高斯变得可见，可能会影响4DGS的对时空变化的感知能力，如(b)中所示。We present \ours, 基于高斯的尺度变化来调整膨胀尺度以减少滤波操作对4DGS的影响，在不降低重建质量的同时保证4DGS的抗锯齿能力，消除采样率变化时出现的伪影。
% \makeatother

\maketitle
\input{sec/0_abstract}    
\input{sec/1_intro}

\input{sec/2_related_work}

\input{sec/3_preliminaries}

\input{sec/4_mothod}

\input{sec/5_experiments}

\input{sec/6_conclusion}
\clearpage
{
    \small
    \bibliographystyle{ieeenat_fullname}
    \bibliography{main}
}

\input{sec/X_suppl}

\end{document}

%% file: preamble.tex
\usepackage{times}
\usepackage{graphicx}
\usepackage{calc}
\usepackage[notransparent]{svg}
\usepackage{amssymb}
\usepackage{amsmath}
\usepackage{xcolor}
\definecolor{iccvblue}{rgb}{0.21,0.49,0.74}
\usepackage[pagebackref,breaklinks,colorlinks,allcolors=iccvblue]{hyperref}

\usepackage{booktabs} % professional-quality tables
\usepackage{multirow} % used by tables
\usepackage{multicol} % used by tables  % clashes with global twocolumn option
\usepackage{tablefootnote}
\usepackage{colortbl} % used by tables
\usepackage{diagbox}  % used in the table with the backslashbox
\usepackage{siunitx}

\usepackage{xspace} % deal with spacing after newcommand
\usepackage[subrefformat=parens,labelformat=parens,font=small]{subcaption} % provides subfigure
\usepackage{wrapfig}

\usepackage{algorithm}
\usepackage{listings}
\newcommand{\ours}{Alias-free-4DGS\xspace}

\newcounter{boldpara}
\newcommand{\boldparagraph}[1]{
    \refstepcounter{boldpara}
    \vspace{0.1em}\noindent{\bf #1}
}

\definecolor{mblue}{HTML}{367dbd}
\colorlet{colorFst}{mblue!45}     % first
\colorlet{colorSnd}{mblue!25}     % second
\colorlet{colorTrd}{yellow!30}      % third
\colorlet{colorLow}{darkgray!30}    % low-light color
\definecolor{R1}{HTML}{E97451}
\definecolor{R2}{HTML}{008080}
\definecolor{R3}{HTML}{0047AB}
\colorlet{cmt}{darkgray!80}         % low-light color
\colorlet{supp}{darkgray!50}        % low-light color

\newcommand{\best}[1]{\cellcolor{red!30} #1}
\newcommand{\second}[1]{\cellcolor{orange!30} #1}
\newcommand{\third}[1]{\cellcolor{yellow!30} #1}

%% file: sec/0_abstract.tex
\begin{abstract}
Existing dynamic scene reconstruction methods based on Gaussian Splatting enable real-time rendering and generate realistic images. However, adjusting the camera's focal length or the distance between Gaussian primitives and the camera to modify rendering resolution often introduces strong artifacts, stemming from the frequency constraints of 4D Gaussians and Gaussian scale mismatch induced by the 2D dilated filter. To address this, we derive a maximum sampling frequency formulation for 4D Gaussian Splatting and introduce a 4D scale-adaptive filter and scale loss, which flexibly regulates the sampling frequency of 4D Gaussian Splatting. Our approach eliminates high-frequency artifacts under increased rendering frequencies while effectively reducing redundant Gaussians in multi-view video reconstruction. We validate the proposed method through monocular and multi-view video reconstruction experiments.Ours project page: \url{https://4d-alias-free.github.io/4D-Alias-free/}

\end{abstract}

%% file: sec/1_intro.tex
\section{Introduction}
\label{sec:intro}

\begin{figure}[t] 
\centering
\includegraphics[width=0.98\columnwidth]{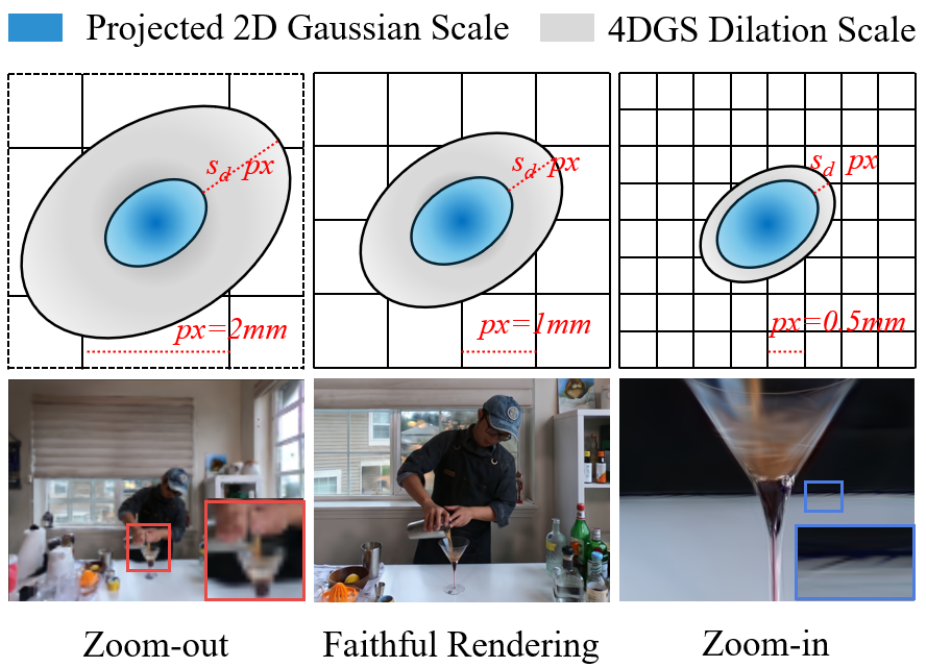}

\caption{ Similar to 3DGS~\cite{3dgs}, 4DGS undergoes substantial dilation when the sampling rate is lowered and experiences erosion along with high-frequency artifacts when the sampling rate is increased.This occurs because changes in resolution alter the effective pixel size while the Gaussian scale remains fixed, causing a mismatch between the Gaussian scale and the filter dilation scale, compounded by the absence of a maximum sampling frequency constraint in 4DGS.}
\label{fig:mis_match}
\vspace{-14pt}
\end{figure}

Reconstructing dynamic scenes from monocular or multi-view videos has garnered significant attention due to its wide applications in augmented reality and virtual reality~\citep{Steuer1992DefiningVR,3dgs,Waisberg2023TheFO,Qu_2024_CVPR,Yan_2024_CVPR,gao2025partrm,wang2025unifying}. In addition to methods based on neural radiance fields (NeRFs) \cite{attal2023hyperreel, li2022neural, li2021neural, park2021hypernerf, fridovich2023k,qu2024livescene,yang2024spectrally,yuan2024slimmerf,chen2023nerrf}, 3D Gaussian Splatting (3DGS) \cite{3dgs,dai20254d,jiang2024gaussianshader} has emerged as a key approach for dynamic scene reconstruction, owing to its ability to render high-quality novel views in real-time. Recent advancements have extended 3DGS to the 4D domain, primarily through two approaches. The first approach employs 4D Gaussians in the spatiotemporal domain~\cite{li2024spacetime,yang2023real, cho20244d}, computing Gaussian attributes across different time frames. The second approach models scene dynamics by deforming Gaussians over time~\cite{bae2024per, lu2025dn, shaw2023swings, 4dGaussians, d3dgs, huang2024sc}. Both methods require the acquisition of 3DGS Gaussian representations at time \( t \) before rendering, with the distinction lying in the way Gaussian distributions over time are learned. We refer to both approaches collectively as 4DGS.

When the camera focal length or object-camera distance changes, altering the sampling rate, 4D Gaussian Splatting (4DGS) exhibits significant artifacts due to Gaussian scale mismatch caused by the 2D dilation filter, a known issue in 3DGS that persists in 4DGS. The 2D dilation filter expands Gaussians in screen space. As rendering resolution changes, pixel size varies while the Gaussian remains fixed, disrupting the dilation-to-Gaussian scale ratio and introducing severe artifacts, as shown in~\cref{fig:mis_match}.
Additionally, existing 4DGS methods lack constraints on Gaussian frequency. Especially in multi-view video reconstruction, modeling object motion with Gaussian movement is challenging, leading to redundant Gaussians. These small redundancies generate high-frequency artifacts when the sampling rate increases. However, incorporating the 3D smoothing filter from 3DGS \cite{mip-splatting} into monocular video reconstruction methods \cite{d3dgs} to constrain 4DGS frequency results in a decline in reconstruction quality. Our analysis shows that Gaussian properties, such as position and scale, change over time in 4DGS. A fixed dilation scale can make imperceptible small Gaussians visible and distort the scale ratio of anisotropic Gaussians, limiting 4DGS’s ability to capture spatio-temporal variations, as shown in~\cref{fig:teaser}.
To address the above issues, we first derive the maximum sampling frequency formula for 4DGS based on the Nyquist-Shannon Sampling Theorem \cite{nyquist1928certain, shannon1949communication}, and propose a more flexible 4D scale-adaptive filter. Our key insights are twofold: first, we avoid dilating overly small Gaussians to prevent invisible Gaussians from becoming visible; second, when a Gaussian’s scale in a given dimension is smaller than or close to the dilation scale, a fixed dilation scale alters the scale ratio between dimensions, leading to anisotropic changes in the Gaussian. The 4D scale-adaptive filter adjusts the dilation scale within a certain range by computing the ratio of Gaussian scales before and after temporal changes, preserving the dimensional proportions after dilation and reducing the impact of filtering on Gaussian anisotropy, as shown in~\cref{fig:teaser}. Additionally, we propose a scale loss to constrain the Gaussian’s own scale, as filtering errors become more significant when the Gaussian scale is smaller than the filter scale. By distributing constraints between the filter and the Gaussian scale, we reduce the minimum dilation scale of the filter. While constraining the maximum frequency of 4DGS, we introduce the 2D Mip filter \cite{mip-splatting} to address the Gaussian scale mismatch in 4DGS.

\noindent  In summary, we make the following contributions:
\begin{itemize}
    \item We propose a maximum sampling frequency calculation method for 4DGS, from which we introduce the 4D scale-adaptive filter and scale loss to flexibly constrain the frequency of 4DGS, while incorporating the 2D Mip filter \cite{mip-splatting} for a general anti-aliasing solution in 4DGS.
    \item We integrate our method into the monocular video reconstruction algorithm D3DGS \cite{d3dgs}, demonstrating the superiority of the 4D scale-adaptive filter, which improves rendering quality at various resolutions without sacrificing full-resolution reconstruction quality.
    \item We integrate our method into the multi-view video reconstruction algorithm 4DGaussian \cite{4dGaussians}, where constraining the maximum sampling frequency of 4DGS effectively reduces redundant Gaussians, improves reconstruction quality, and eliminates high-frequency artifacts when increasing the sampling rate, as shown in \cref{fig:dynerf_cmp1}.
\end{itemize}

%% file: sec/2_related_work.tex
\begin{figure}[t] 
\centering
\includegraphics[width=1.0\columnwidth]{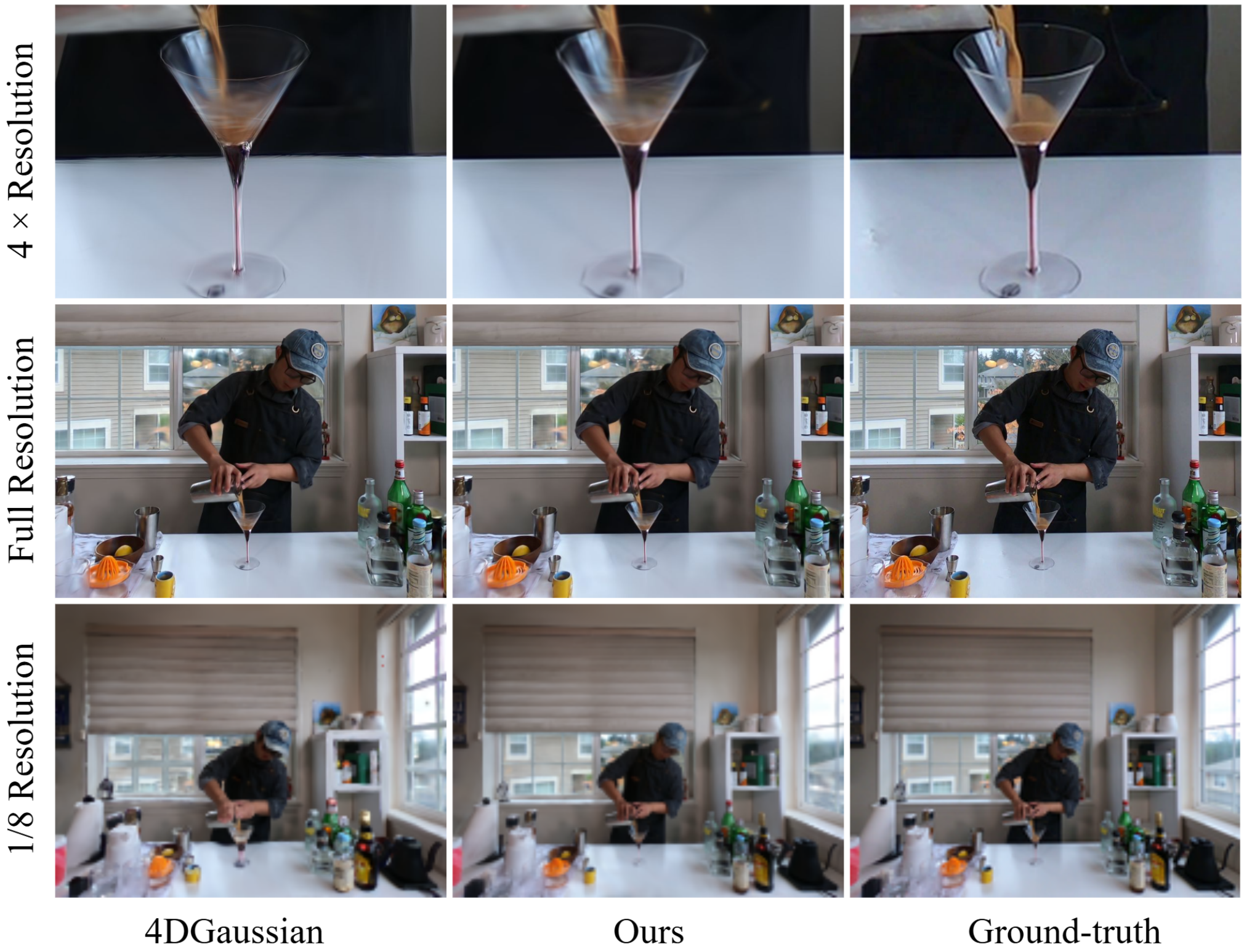} 
\caption{We trained all models on single-scale images (full resolution) and rendered images at different resolutions by adjusting the focal length. When the sampling rate changes, 4D Gaussians\cite{4dGaussians} exhibit strong artifacts, which our method effectively eliminates.}
\label{fig:dynerf_cmp1}
\vspace{-14pt}
\end{figure}

% Mip-Splatting~\cite{mip-splatting}提出的3D Smoothing Filter固定的膨胀尺度会让4DGS中尺度极小的不可见高斯变得可见，同时可能会使高斯不同维度间的尺度比例发生变化，影响高斯的各向异性。We present \ours，通关使用4D Scale-adaptive Filter和Scale Regularization联合约束4DGS的最大采样频率来减小滤波器的最小膨胀尺度，同时避免了高斯尺度远小于滤波器膨胀尺度时滤波操作会产生较大误差。此外，我们提出的4D Scale-adaptive Filter通过掩码避免膨胀不可见高斯以及通过用当前时间帧下高斯变化前后的尺度比例来调整滤波膨胀尺度在一定程度下减小滤波器对高斯各向异性的影响。\ours在不影响重建质量的同时能够有效消除采样频率变化时出现的伪影。

\begin{figure*}[t] % [t] 表示尽量放在页面顶部
  \centering
  \includegraphics[width=1.0\textwidth]{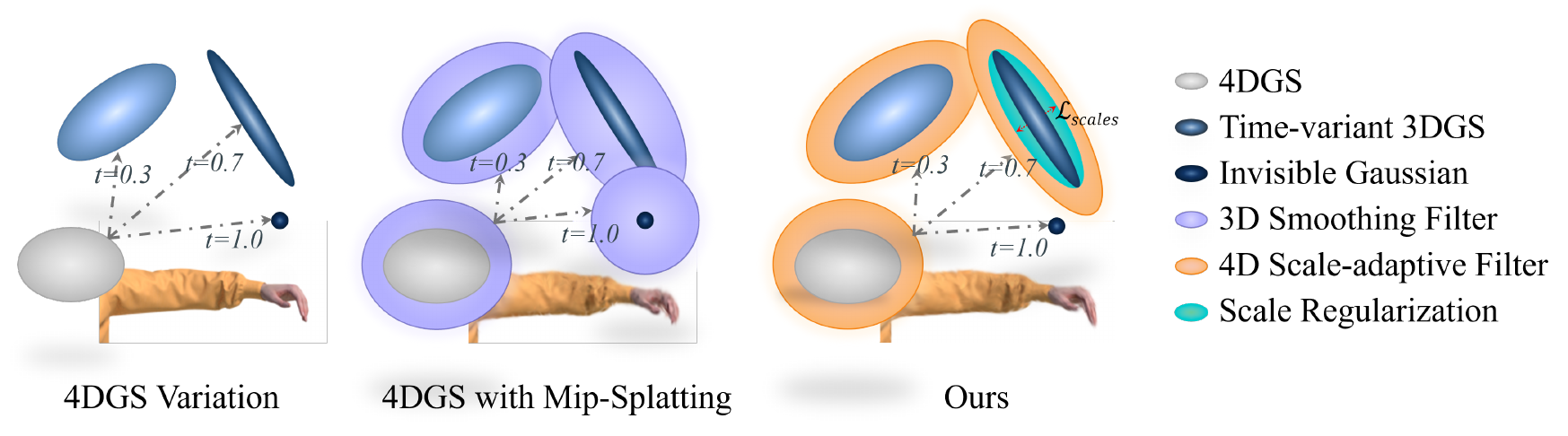}
  %\captionsetup{type=figure}
  \caption{
    The fixed dilation scale used by the 3D smoothing filter proposed in Mip-Splatting~\cite{mip-splatting} can inadvertently render imperceptible Gaussians with extremely small scales visible in 4DGS, and may alter the scale ratios among Gaussian dimensions, affecting Gaussian anisotropy. We present \ours, which leverages a 4D Scale-adaptive Filter and Scale Regularization to jointly constrain the maximum sampling frequency of 4DGS. This reduces the filter’s minimum dilation scale and avoids significant filtering errors when Gaussian scales are much smaller than the filter’s dilation scale. Moreover, our 4D Scale-adaptive Filter masks out imperceptible Gaussians and adapts the dilation scale using the ratio of Gaussian scales before and after changes in the current time frame, thus mitigating the filter’s impact on Gaussian anisotropy. \ours effectively eliminates artifacts that arise from varying sampling frequencies without compromising reconstruction quality.}
  
    % The fixed expansion scale in the 3DGS anti-aliasing algorithm Mip-Splatting~\cite{mip-splatting} influences the temporal anisotropy of Gaussians and can render otherwise invisible Gaussians visible, potentially impairing the spatiotemporal perception of 4DGS. We present \ours, which adjusts the dilation scale based on Gaussian scale variations to mitigate the impact of filtering on 4DGS, while preserving antialiasing capability and maintaining reconstruction quality, effectively eliminating artifacts caused by sampling rate changes.}
  \label{fig:teaser}
  \vspace{-10pt}
\end{figure*}

\section{Related Works}
\textbf{Dynamic 3D Gaussians:} In recent years, 3D Gaussian Splatting (3DGS)~\cite{3dgs,li20253d,yang2024spectrally,li2025mipmap} has achieved remarkable progress in novel view synthesis, enabling real-time rendering at high-definition resolutions. Unlike ray tracing, 3DGS explicitly represents a scene as a set of 3D Gaussians and utilizes rasterization for rendering. Recent advancements have extended 3DGS to 4D, primarily following two approaches. The first employs 4D Gaussians in the spatiotemporal domain~\cite{li2024spacetime, yang2023real, cho20244d}, computing Gaussian attributes across different time frames. This approach achieves high visual quality and real-time performance in multi-view video reconstruction. However, its limited accuracy in modeling Gaussian motion introduces significant redundancy, making it less effective for monocular reconstruction. The second approach models scene dynamics by deforming Gaussians over time~\cite{bae2024per, lu2025dn, shaw2023swings, 4dGaussians, d3dgs, huang2024sc,chen2024freegaussian}, enabling precise motion estimation while maintaining a compact representation. This method is well-suited for both monocular and multi-view video reconstruction.

\noindent \textbf{Anti-Aliasing in Neural Rendering:} Aliasing is a common issue in computer graphics, arising when rendering frequency changes abruptly, leading to visual artifacts. Existing anti-aliasing techniques can be categorized into super-sampling and pre-filtering. The former increases the sampling rate to reconstruct high-frequency details in the scene~\cite{akeley1993reality, deering1988triangle, fuchs1985fast, haeberli1990accumulation, mammen1989transparency, whitted2005improved}, while the latter mitigates aliasing by applying filters to suppress high-frequency components~\cite{crow1984summed, heckbert1989fundamentals, mueller1998splatting, swan1997anti, williams1983pyramidal, zwicker2001ewa}. In neural rendering, NeRF-based methods~\cite{barron2021mip, barron2023zip, hu2023tri,liu2024rip} achieve effective anti-aliasing in static scenes, while DMiT extends pre-filtering to dynamic NeRFs, enabling aliasing suppression in dynamic scenes. 3DGS~\cite{3dgs} suffers from severe artifacts when the dilation scale of the 2D dilation filter deviates from the intrinsic scale of Gaussian primitives during rendering. SA-GS~\cite{sa-gs} addresses this issue with a scale-adaptive 2D filter and integrates super-sampling but lacks robust anti-aliasing when increasing the sampling rate. Mip-Splatting~\cite{mip-splatting} employs a hybrid filtering mechanism to suppress high-frequency components in both 2D and 3D Gaussians, achieving anti-aliasing. However, its fixed dilation scale limits flexibility, reducing its ability to preserve fine details and capture spatiotemporal variations in 4DGS, ultimately degrading reconstruction quality.

%% file: sec/3_preliminaries.tex
\section{Preliminaries}
\label{sec:Preliminaries}
\noindent \textbf{3D Gaussian Splatting:}
Previous work \cite{kerbl20233d, zwicker2001ewa} proposed representing a 3D scene as a collection of scaled 3D Gaussian primitives and using volume splatting for image rendering. The geometry of each Gaussian primitive is defined in world coordinates by its opacity $\alpha_k \in [0, 1]$, center position $\mathbf{p}_k \in \mathbb{R}^{3 \times 1}$, and covariance matrix $\mathbf{\Sigma}_k \in \mathbb{R}^{3 \times 3}$:
\begin{equation} 
\mathcal{G}_k (\mathbf{x}) = e^{-\frac{1}{2} (\mathbf{x} - \mathbf{p}_k)^T \mathbf{\Sigma}_k^{-1} (\mathbf{x} - \mathbf{p}_k)} \end{equation}
For optimization, the covariance matrix $\Sigma_k$ is decomposed as $\Sigma_k = R S S^T R^T$, where $R$ is a rotation matrix represented by a quaternion $r \in SO(3)$, and $S$ is a scaling matrix represented by a 3D vector $\mathbf{s}$. The 3D Gaussians can be projected onto 2D and rendered for each pixel using the following 2D covariance matrix $\Sigma'_k$:
\begin{equation}
\Sigma_k' = J V \Sigma_k V^T J^T
\end{equation}
Here, $J$ denotes the Jacobian of the affine approximation of the projective transformation, and $V$ represents the view matrix, which transitions coordinates from world space to camera space. By omitting the third row and column of $\Sigma'_k$, we obtain a 2D covariance matrix $\Sigma^{2D}_k$ in ray space. We use $G^{2D}_k$ to refer to the corresponding scaled 2D Gaussian.
Finally, 3DGS \cite{kerbl20233d} uses spherical harmonics to model the view-dependent color $c_k$ and renders the image via alpha blending according to the depth order of the primitives, which is $1, \dots, K$:
\begin{equation} 
c(x) = \sum_{k=1}^{K} c_k \alpha_k G^{2D}k(x) \prod_{j=1}^{k-1} \left( 1 - \alpha_j G^{2D}_j (x) \right) 
\end{equation}

\noindent \textbf{4D Gaussian Splatting:}
The common approach for modeling dynamic scenes in both monocular and multi-view videos \cite{d3dgs,4dGaussians} involves decoupling 4D Gaussians into 3D Gaussians and a deformation field. During the warm-up phase, a set of static Gaussians $G(p, r, s, \alpha)$ is first trained. Subsequently, the deformation field learns the changes in position $\Delta p$, rotation $\Delta r$, and scale $\Delta s$ based on the Gaussian’s position $p$ over time $t$. The deformed 3D Gaussians $G(p + \Delta p, r + \Delta r, s + \Delta s, \alpha)$ are then used for rendering. The 3D Gaussians and deformation network are optimized jointly through the fast backward pass by tracking accumulated $\alpha$ values, together with the adaptive control of the Gaussian density.

\noindent \textbf{Filter:}
To avoid issues such as 2D Gaussians being smaller than one pixel after projection onto screen space, 3DGS employs a 2D dilation filter for low-pass filtering:
\begin{equation}
G_k^{2D}(x) = e^{-\frac{1}{2} (x - p_k)^T (\Sigma_k^{2D} + \sigma_sI)^{-1} (x - p_k)}
\end{equation}
where I is a 2D identity matrix and $\sigma_s$ is a scalar dilation hyperparameter. Mip-Splatting\cite{mip-splatting} replaces the 2D dilation filter with a 2D mip filter, effectively mitigating aliasing and dilation issues:
\begin{equation}
G_k^{2D}(x)_{mip} = \sqrt{\frac{|\Sigma_k^{2D}|}{|\Sigma_k^{2D} + \sigma_sI|}} e^{-\frac{1}{2} (x - p_k)^T (\Sigma_k^{2D} + \sigma_sI)^{-1} (x - p_k)}
\end{equation}
Additionally, in MIP-splatting, a 3D smoothing filter is applied to limit the maximum sampling frequency of the Gaussians, thereby eliminating high-frequency artifacts that arise when increasing the sampling rate:
\begin{equation}
\label{eq:3d_smoothing_filter}
G_k(x)_{reg} = \sqrt{\frac{|\Sigma_k|}{|\Sigma_k + \frac{\sigma_s}{{\hat{\nu}_k}^2} \cdot I|}} e^{-\frac{1}{2} (x - p_k)^T (\Sigma_k + \frac{\sigma_s}{{\hat{\nu}_k}^2} \cdot I)^{-1} (x - p_k)}
\end{equation}
$\hat{\nu}_k$ is the maximal sampling rate for primitive k.

\begin{table*}[htb]
  \centering
  \resizebox{\textwidth}{!}{%
  \begin{tabular}{l|ccccc|ccccc|ccccc}
  \hline
  \multirow{2}{*}{\textbf{Methods}} 
  & \multicolumn{5}{c|}{\textbf{PSNR $\uparrow$}} 
  & \multicolumn{5}{c|}{\textbf{SSIM $\uparrow$}} 
  & \multicolumn{5}{c}{\textbf{LPIPS$_V$ $\downarrow$}} \\
  \cline{2-6} \cline{7-11} \cline{12-16}
   & Full Res. & 1/2 Res. & 1/4 Res. & 1/8 Res. & Avg. 
   & Full Res. & 1/2 Res. & 1/4 Res. & 1/8 Res. & Avg. 
   & Full Res. & 1/2 Res. & 1/4 Res. & 1/8 Res. & Avg. \\
  \hline
  D-NeRF \cite{dnerf}
  & 28.01 & 28.37 & 29.49 & 28.94 & 28.70 
  & 0.935 & 0.947 & 0.951 & 0.947 & 0.944 
  & 0.065 & 0.052 & 0.063 & 0.064 & 0.062 \\
  TiNeuVox \cite{fang2022fast} 
  & 31.24 & 32.12 & 32.55 & 30.50 & 31.60 
  & 0.962 & 0.969 & 0.975 & 0.966 & 0.968 
  & 0.059 & 0.045 & 0.035 & 0.046 & 0.046 \\
  K-Planes-H \cite{fridovich2023k} 
  & 27.26 & 27.67 & 27.97 & 27.92 & 27.61 
  & 0.955 & 0.954 & 0.953 & 0.949 & 0.952 
  & 0.062 & 0.056 & 0.054 & 0.063 & 0.059 \\
  K-Planes-E \cite{fridovich2023k} 
  & 26.80 & 27.18 & 27.58 & 27.34 & 27.23 
  & 0.951 & 0.949 & 0.948 & 0.941 & 0.947 
  & 0.069 & 0.065 & 0.064 & 0.068 & 0.067 \\
  Tensor4D \cite{shao2023tensor4d}  
  & 25.25 & 25.67 & 26.04 & 25.15 & 25.49 
  & 0.932 & 0.934 & 0.938 & 0.921 & 0.931 
  & 0.101 & 0.093 & 0.071 & 0.050 & 0.079 \\
  4DGS \cite{4dGaussians} 
  & 30.13 & 30.36 & 30.84 & 30.63 & 30.49 
  & 0.963 & 0.966 & 0.968 & 0.967 & 0.966 
  & 0.048 & 0.042 & 0.042 & 0.038 & 0.042 \\
  D3DGS \cite{d3dgs} 
  & \cellcolor{yellow!30} 32.40 & \cellcolor{yellow!30} 32.60 & \cellcolor{yellow!30} 32.96 & \cellcolor{yellow!30} 32.76 & \cellcolor{yellow!30} 32.68 
  & \cellcolor{yellow!30} 0.976 & \cellcolor{yellow!30} 0.978 & \cellcolor{yellow!30} 0.979 & \cellcolor{yellow!30} 0.979 & \cellcolor{yellow!30} 0.978 
  & \cellcolor{yellow!30} 0.032 & \cellcolor{yellow!30} 0.027 & \cellcolor{yellow!30} 0.027 & \cellcolor{yellow!30} 0.025 & \cellcolor{yellow!30} 0.028 \\
  DMiT \cite{dmit}
  & \cellcolor{orange!30} 34.15 & \cellcolor{orange!30} 35.04 & \cellcolor{orange!30} 35.81 & \cellcolor{orange!30} 36.09 & \cellcolor{orange!30} 35.27 
  & \cellcolor{orange!30} 0.980 & \cellcolor{orange!30} 0.984 & \cellcolor{orange!30} 0.987 & \cellcolor{orange!30} 0.988 & \cellcolor{orange!30} 0.985
  & \cellcolor{orange!30} 0.029 & \cellcolor{orange!30} 0.020 & \cellcolor{orange!30} 0.014 & \cellcolor{orange!30} 0.010 & \cellcolor{orange!30} 0.019 \\
  \hline
  Ours  
  & \cellcolor{red!30} 36.99 & \cellcolor{red!30} 38.72	& \cellcolor{red!30} 39.61	& \cellcolor{red!30} 38.96 & \cellcolor{red!30} 38.57
  & \cellcolor{red!30} 0.982 & \cellcolor{red!30} 0.987	& \cellcolor{red!30} 0.99	& \cellcolor{red!30} 0.992  & \cellcolor{red!30} 0.988
  & \cellcolor{red!30} 0.025 & \cellcolor{red!30} 0.012	& \cellcolor{red!30} 0.008 & \cellcolor{red!30} 0.008 & \cellcolor{red!30} 0.013 \\
  \hline
  \end{tabular}%
  }
  \caption{Multi-scale Training and Multi-scale Testing on the D-NeRF dataset~\cite{dnerf}. Our approach achieves state-of-the-art performance in most metrics.}
  \label{tab:dnerf_multi_train}
\end{table*}

%% file: sec/4_mothod.tex
\section{Methods}
% 高频约束，高斯模糊、降低膨胀对重建的影响
% 主要卖点是在不影响重建质量的前提提高抗锯齿能力
\subsection{Max Sampling for 4D Gaussian}
Given the distance $d$ between the Gaussian element and the camera, and the camera's focal length $f$, it is straightforward to compute the sampling interval $\hat{T}$ and the sampling frequency $\hat{\nu}$ of the Gaussian element:
\begin{equation}
\hat{T} = \frac{1}{\hat{\nu}} = \frac{d}{f}
\end{equation}
To determine the maximum sampling frequency $\hat{\nu}$ for a single Gaussian element $k$, we minimize the ratio $\frac{d}{f}$. Let \( N \) denote the total number of cameras, the maximum sampling frequency can be determined as:
\begin{equation}
\label{eq:max_sample}
\frac{1}{\hat{\nu}_k} = \min_{n \in \{1, \dots, N\}} \left( \left\{ \mathds{1}_n(p_k(t)) \cdot \frac{d_n(t)}{f_n(t)} \right\}_{t=1}^{T} \right)
\end{equation}
Typically, \( f_n(t) \) is fixed at a constant interval, and \( p_k(t) = p_k + \Delta p_k(t) \). We approximate the depth \( d_n(t) \) using the center of the primitive \( p_k(t) \). The function \( \mathds{1}_n(p_k(t)) \) acts as an indicator function that assesses the visibility of a primitive. Equation \cref{eq:max_sample} provides a clear and intuitive formulation for calculating the maximum sampling frequency of the 4D Gaussian. However, the computational cost associated with evaluating \( \Delta p_k(t) \) within the 4D Gaussian framework is substantial. When \( T \) is large, recalculating the maximum sampling frequency by iterating over all \( \Delta p_k(t) \) values at each step can significantly extend the training duration. Consequently, we propose an approximate method to compute the maximum sampling frequency of the 4D Gaussian more efficiently.

During the warm-up phase and the early stage of training the deformation field, we use the initial positions of the Gaussian basis functions \( p_k \) to compute the rough minimum sampling interval \( \hat{T}_k \):
\begin{equation}
\hat{T}_k = \min \left( \left\{ 1_n(p_k) \cdot \frac{d_n}{f_n} \right\}_{n=1}^{N} \right)
\label{eq:static_max_t}
\end{equation}
Once the deformation field stabilizes, we employ a momentum-based update method to learn a more accurate sampling interval \( \hat{T}_k \):
\begin{equation}
    \label{eq:min_T}
  \hat{T}_k = \left( 1 - \lambda_v \right)   \cdot \hat{T}_k + \lambda_v \cdot \min\left( \hat{T}_k , \frac{d_n(t)}{f_n(t)} \right)
\end{equation}
$\hat{\nu}_k$ is the reciprocal of $\hat{T}_k$. Note that although \( d_n(t) \) still requires the computation of \( \Delta p_k(t) \), due to the need to iterate over calculations in \cref{eq:min_T}, we can directly get \( d_n(t) \) during training in CUDA, as \( d_n(t) \) is computed during the CUDA rendering process, making the overhead of \cref{eq:min_T} negligible.

\begin{table*}[htb]
  \centering
  \resizebox{\textwidth}{!}{%
  \begin{tabular}{l|ccccc|ccccc|ccccc}
  \hline
  \multirow{2}{*}{\textbf{Methods}} 
  & \multicolumn{5}{c|}{\textbf{PSNR $ \uparrow $}} 
  & \multicolumn{5}{c|}{\textbf{SSIM $ \uparrow $ }} 
  & \multicolumn{5}{c}{\textbf{LPIPS$_v$ $ \downarrow $}} \\
  \cline{2-6} \cline{7-11} \cline{12-16}
   & Full Res. & 1/2 Res. & 1/4 Res. & 1/8 Res. & Avg. 
   & Full Res. & 1/2 Res. & 1/4 Res. & 1/8 Res. & Avg. 
   & Full Res. & 1/2 Res. & 1/4 Res. & 1/8 Res. & Avg. \\
  \hline
  D3DGS \cite{d3dgs}* 
  & \cellcolor{yellow!30} 38.30	& 32.48	& 26.63	& 22.35	& 29.94 
  & \cellcolor{red!30} 0.986	& 0.975	& 0.932	& 0.857	& 0.937
  & \cellcolor{red!30} 0.017	& 0.016	& 0.035	& 0.072	& 0.035 \\
  D3DGS \cite{d3dgs} - Dilation
  & 35.45	& 35.64	& 34.07	& 31.47	& 34.16 
  & 0.977	& 0.980	& 0.979	& 0.973	& 0.977
  & 0.030	& 0.022	& 0.028	& 0.054	& 0.034 \\
  2D Mip Filter \cite{mip-splatting}
  & \cellcolor{red!30} 38.45	& \cellcolor{orange!30} 38.20	& \cellcolor{yellow!30} 35.61	& \cellcolor{yellow!30} 32.42	& \cellcolor{yellow!30} 36.17 
  & \cellcolor{orange!30} 0.985	& \cellcolor{red!30} 0.988	& \cellcolor{orange!30} 0.986	& \cellcolor{yellow!30} 0.980  & \cellcolor{orange!30} 0.985
  & \cellcolor{orange!30} 0.018	& \cellcolor{red!30} 0.011 & \cellcolor{orange!30} 0.012	& \cellcolor{orange!30}0.019	& \cellcolor{orange!30} 0.015 \\
  Mip-Splatting\textsubscript{4D} \cite{mip-splatting}
  & 37.76	& \cellcolor{yellow!30} 38.02	& \best 36.15	& \cellcolor{orange!30} 33.29	& \cellcolor{orange!30} 36.31
  & 0.984	& \cellcolor{orange!30} 0.987	& \cellcolor{red!30} 0.987	& \cellcolor{orange!30} 0.983	& \cellcolor{orange!30} 0.985
  & 0.020	& \cellcolor{red!30} 0.011	& \cellcolor{red!30} 0.011	& \second 0.017	& \cellcolor{orange!30} 0.015 \\
  \hline
  Ours 
  & \cellcolor{orange!30} 38.39	& \cellcolor{red!30} 38.40	& \second 35.93	& \cellcolor{red!30} 33.81	& \cellcolor{red!30} 36.64
  & \cellcolor{orange!30} 0.985	& \cellcolor{red!30} 0.988	& \cellcolor{red!30} 0.987	& \cellcolor{red!30} 0.984	& \cellcolor{red!30} 0.986
  & \cellcolor{red!30} 0.017	& \cellcolor{red!30} 0.011	& \best 0.011	&\cellcolor{red!30}  0.016	& \cellcolor{red!30} 0.014 \\
  \hline
  \end{tabular} %
  }
  \caption{Single-scale Training and Multi-scale Testing on the D-NeRF dataset~\cite{dnerf}. All methods are trained on full-resolution images and evaluated at full resolution and three progressively lower resolutions, simulating zoom-out effects. Our approach preserves high-quality rendering at reduced resolutions while maintaining reconstruction fidelity at full resolution. * denotes models that we retrained. } 
  \label{tab:dnerf_ds}
\end{table*}

\subsection{4D Scale-adaptive Filter}
%这部分先写的比较细，后面会把部分内容放到intro去
We introduce the 4D Scale-adaptive Filter at the maximum sampling frequency \( \hat{\nu}_k \) of the 4D Gaussian primitive \( k \). This approach is based on two considerations: First, the 4D Gaussian primitive \( k \) may be invisible at a specific time \( t \), necessitating the avoidance of dilation at that time. Second, the 4D Gaussian models variations in scene information through scale changes \( \Delta s \). When \( s_t = s + \Delta s(t) \) in a given dimension becomes smaller than or approaches the dilation factor \( \frac{\sqrt{\sigma_s}}{\hat{\nu}_k} \), fixed-scale dilation can distort the scale ratio between dimensions. The 4D Scale-adaptive Filter primarily adjusts the dilation coefficient based on scale variations, thereby flexibly constraining the frequency of the 4D Gaussian. First, the proportional change in the dilation coefficient,$\rho_{\text{adapt}}$, is computed according to the scale variation ratio:
% The 4D Scale-adaptive Filter achieves flexible constraints on the 4D Scale-adaptive frequency by adjusting \( \sigma_\text{adapt} \):
% \begin{align}
% \rho_{\text{adapt}} = \operatorname{clip} \left( \frac{s_t^2}{s^2}, \; \frac{ \rho_\text{min} }{max((\Delta R_p)^2, 1)}, \; \rho_\text{max} \right)
% \end{align}
\begin{align}
\rho_{\text{adapt}} = \operatorname{clip}\left(
    \frac{s_t^2}{s^2},\quad
    \rho_{\text{min}},\quad
    \rho_{\text{max}}
\right)
\end{align}
The dilation coefficient is then updated for the visible Gaussian using $\rho_{\text{adapt}}$. The minimum dilation ratio $\rho_{\text{min}}$ constrains the Gaussian's maximum frequency, while the maximum dilation ratio $\rho_{\text{max}}$ expands the dilation scale of enlarged dimensions within a certain range to preserve the dimensional scale ratio before and after dilation, mitigating the impact of filtering on Gaussian anisotropy.
\begin{align}
\sigma_{\text{adapt}} = 
\begin{cases} 
\rho_{\text{adapt}} \cdot \sigma_s   & \text{if } s_t^2 \geq \rho_{\text{thre}} \cdot \frac{\sigma_s}{\hat{\nu}_k^2}, \\
\epsilon \cdot \sigma_s  & \text{otherwise}.
\end{cases}
\label{eq:mask}
\end{align}
To prevent imperceptible Gaussians from becoming visible, we assign a minimal dilation scale $\epsilon$ to Gaussians with scales below a threshold. Filtering is then applied based on $\sigma_{\text{adapt}}$:
\begin{align} %格式之后再调
G_k(x, t)_{4D} = & \; \sqrt{\frac{|\Sigma_k(t)|}{|\Sigma_k(t) + \frac{\sigma_{\text{adapt}}}{{\hat{\nu}_k}^2} \cdot I|}} \cdot \nonumber \\
& \; e^{-\frac{1}{2} (x - p_k(t))^T \left( \Sigma_k(t) + \frac{\sigma_{\text{adapt}}}{{\hat{\nu}_k}^2} \cdot I \right)^{-1} (x - p_k(t))}
\end{align}
% The 4D Scale-adaptive Filter computes \( \rho_{\text{adapt}} \) using a clipping operation to restrict the dilation coefficient within a specified range. \( \rho_\text{min} \) is used to constrain the maximum frequency of the visible Gaussian, while \( \rho_\text{max} \) is employed to increase the dilation coefficient for the dimensions where \( s + \Delta s(t) \) grows, thereby reducing the filter's impact on the Gaussian's anisotropy. 

\subsection{Scale Regularization}
Our intuition is to minimize the filter’s dilation scale by distributing it between the filter and the Gaussian’s scale constraint. We introduce the scale loss to jointly regulate the maximum sampling frequency of the 4D Gaussian using the 4D Scale-Adaptive Filter:
\begin{align}
\label{eq:scales_loss}
\mathcal{L}_{\text{scales}} = \sum_{k} \left( \left( \rho_{\text{min}} \frac{\sigma_s}{{\hat{\nu}_k}^2} - {s_k(t)}^2 \right) M_1 \right)
\end{align}
where \( s_k(t) = s_k + \Delta s_k(t) \) represents the scale of primitive \( k \) at time \( t \). Here, \( M_1 \) is an indicator function that activates when the scale satisfies \( \rho_{\text{thre}} \cdot \frac{\sigma_s}{{\hat{\nu}_k}^2} < s_k(t)^2 < \rho_{\text{min}} \cdot \frac{\sigma_s}{{\hat{\nu}_k}^2} \).

\subsection{Optimization}
Similar to \cite{d3dgs,4dGaussians}, we first optimize the static Gaussians during the initial 3000 iterations as a warm-up phase, and then jointly optimize both the Gaussians and the deformation field. The training process is supervised by a combination of color loss and scale loss, formulated as:
\begin{align}
    \mathcal{L} = \mathcal{L}_\text{color} + \lambda_1 \cdot \mathcal{L}_{\text{scales}}
\end{align}
where $\mathcal{L}_\text{color}$ aligns with the coda base that integrates our filtering mechanism.

%% file: sec/5_experiments.tex
\begin{figure*}[hbt] 
\centering
\includegraphics[width=0.82\textwidth]{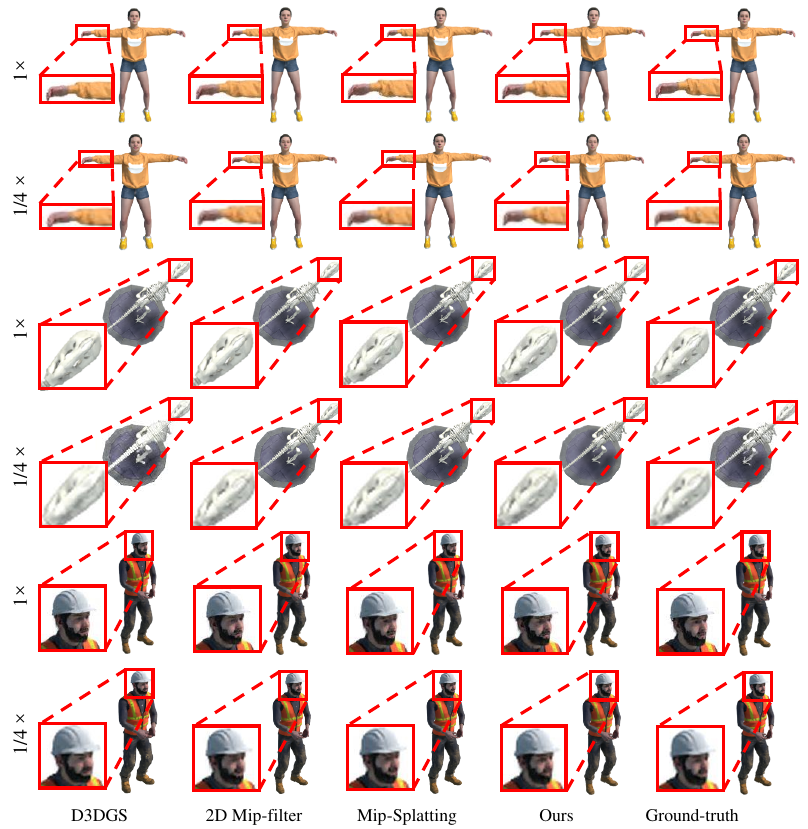} 
\caption{Single-scale Training and Multi-scale Testing on the D-NeRF dataset~\cite{dnerf}.All methods are trained at full resolution and evaluated at various lower resolutions to simulate zoom-out effects. D3DGS~\cite{d3dgs} exhibits noticeable blurring and inflation artifacts at lower resolutions. Integrating Mip-splatting~\cite{mip-splatting} mitigates inflation but introduces local reconstruction distortions, such as the deformation observed in the hand region (first row). In contrast, our method preserves the reconstruction quality of D3DGS while maintaining more realistic visual fidelity at lower resolutions.}
\label{fig:dnerf_ds1}
\end{figure*}

\begin{table*}[h]
\centering
\renewcommand{\arraystretch}{1.2} % Adjust row spacing
\setlength{\tabcolsep}{5pt} % Adjust column spacing for better fit
\resizebox{\textwidth}{!}{ % Ensure table fits within text width
\begin{tabular}{|l|ccc|c|ccc|c|ccc|c|c|}
\hline
\multirow{2}{*}{} 
& \multicolumn{4}{c|}{\textbf{PSNR} $\uparrow$} 
& \multicolumn{4}{c|}{\textbf{SSIM} $\uparrow$} 
& \multicolumn{4}{c|}{\textbf{LPIPS} $\downarrow$} 
& \multirow{2}{*}{$|\mathcal{G}|$} \\
\cline{2-13}
 & 1× Res. & 2× Res. & 4× Res. & Avg. 
 & 1× Res. & 2× Res. & 4× Res. & Avg. 
 & 1× Res. & 2× Res. & 4× Res. & Avg. &  \\
\hline
4DGaussian \cite{4dGaussians}         
& 31.87 & 26.52 & 23.82 & 27.40 
& 0.941 & 0.849 & 0.793 & 0.861
& 0.066 & 0.196 & 0.296 & 0.186 
& 106K \\
2D Mip Filter \cite{mip-splatting}         
& \third 32.79 & \third 29.63 & \third 27.94 & \third 30.12 
& \third 0.964 & \third 0.904 & \third 0.849 & \third 0.906 
& \third 0.041 & \third 0.140 & \third 0.248 & \third 0.143 
& \third 74K \\
Mip-Splatting\textsubscript{4D} \cite{mip-splatting}            
& \best 33.47 & \best 30.51 & \best 29.31 & \best 31.10  
& \best 0.966 & \best 0.916 & \second 0.880 & \best 0.921  
& \best 0.040 & \best 0.138 & \best 0.237 & \best 0.138 
& \second 68K \\  				
\hline
Ours          
&  \second 33.30 &  \second 30.45 &  \second 29.27 &  \second 31.01  
&  \best 0.966 &  \best 0.916 &\best  0.881 &  \best 0.921  
&  \best 0.040 &  \best 0.138 & \second 0.238 & \second 0.139  
& \best 67K \\  
\hline
\end{tabular}
}
\caption{\textbf{Single-Scale Training and Multi-Scale Testing on the N3DV Dataset}~\cite{n3dv}. All methods are trained at 1× and evaluated at 1×, 2×, and 4×, with higher sampling rates simulating zoom-in effects. While our method slightly underperforms Mip-Splatting~\cite{mip-splatting} at the training resolution due to reduced lighting sensitivity, it achieves comparable or superior results at higher sampling rates. Additionally, constraining the frequency of 4DGS reduces the number of Gaussians $|\mathcal{G}|$ and enhances reconstruction quality.}
\label{tab:dynerf_up}
\end{table*}

\begin{figure*}[hbtp] 
\centering
\includegraphics[width=0.92\textwidth]{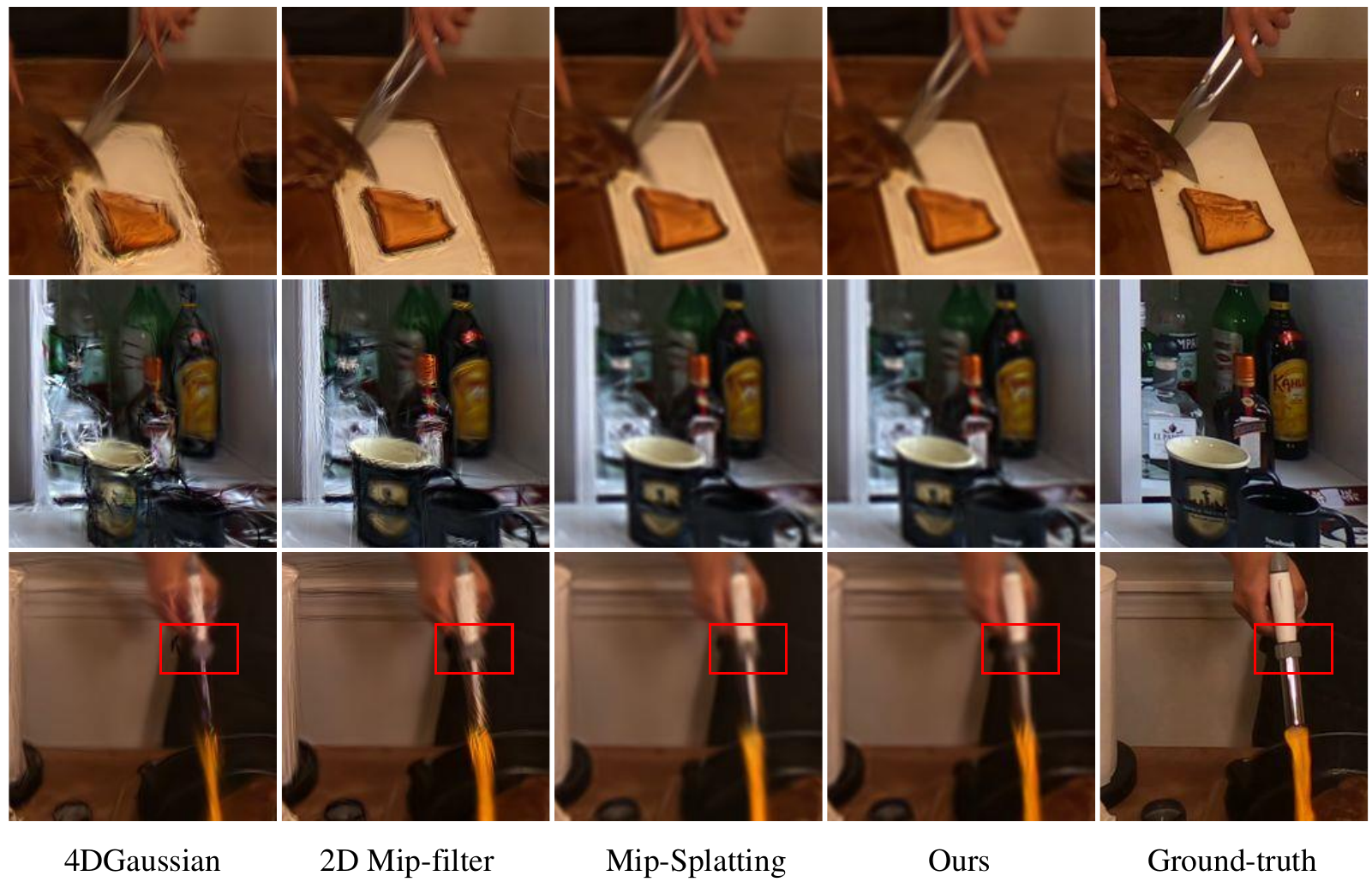} 
\caption{\textbf{Single-Scale Training and Multi-Scale Testing on the N3DV Dataset}~\cite{n3dv}. All models are trained on images downsampled by a factor of four and rendered at full resolution to simulate zoom-in and moving-closer effects. Our method effectively eliminates high-frequency artifacts and produces more complete object shapes compared to Mip-Splatting.}.
\label{fig:dynerf_up1}
\end{figure*}

\begin{table*}[htbp]
\centering
\renewcommand{\arraystretch}{1.2} % 调整行间距
\setlength{\tabcolsep}{4pt} % 调整列间距以适应双栏
\resizebox{0.92\textwidth}{!}{ % 限制表格宽度不超过双栏
\begin{tabular}{|l|ccc|c|ccc|c|ccc|c|}
\hline
\multirow{2}{*}{} & \multicolumn{4}{c|}{\textbf{PSNR} $\uparrow$} & \multicolumn{4}{c|}{\textbf{SSIM} $\uparrow$} & \multicolumn{4}{c|}{\textbf{LPIPS} $\downarrow$} \\
\cline{2-13}
 & 1× Res. & 2× Res. & 4× Res. & Avg. & 1× Res. & 2× Res. & 4× Res. & Avg. & 1× Res. & 2× Res. & 4× Res. & Avg. \\
\hline
D3DGS \cite{d3dgs}*         
& 39.69 & 30.15 & 27.37 & 32.40   
& \second 0.990 & 0.959 & 0.936 & 0.962    								
& \best 0.008 & 0.033 & 0.061 & 0.034 \\
2D Mip Filter \cite{mip-splatting}         
& \second 40.568 & 36.104 & 33.827 & \third 36.833  				
& \second 0.990 & 0.981 & 0.967 & 0.979  							
& \best 0.008 & 0.025 & \best 0.048 & \second 0.027 \\
Mip-Splatting\textsubscript{4D} \cite{mip-splatting}            
& 40.29 & \second 36.26 & \second 34.02 & \second 36.86   				
& \second 0.990 & \best 0.982 & \best 0.969 & \second 0.98  	
& \best 0.008 & \second 0.024 & 0.049 & \second 0.027 \\  				
\hline
Ours w/o $\mathcal{L}_{\text{scales}}$               
& \third 40.402 & \third 36.124 & \third 33.872 & 36.80  												
& \second 0.990 & \best 0.982 & \second 0.968 & \second 0.98 
& \best 0.008 & \second 0.024 & \best 0.048 & \second 0.027 \\

Ours          
& \best 40.61 & \best 36.36 & \best 34.07 & \best 37.01  				
& \best 0.991 & \best 0.982 & \best 0.969 & \best 0.981     				
& \best 0.008 &\best  0.023 & \best  0.048 & \best 0.026 \\  
\hline
\end{tabular}
}
\caption{Single-scale Training and Multi-scale Testing on the D-NeRF dataset~\cite{dnerf}.All methods are trained on the smallest scale (1×) and evaluated across three scales (1×, 2×, and 4x), with evaluations at higher sampling rates simulating zoom-in effects.* denotes models that we retrained.}
\label{tab:dner_up}
\end{table*}

\begin{table}[tb]
\centering
\resizebox{\columnwidth}{!} {
\begin{tabular}{l|c|c|c}
\hline
Method & PSNR $\uparrow$ & SSIM $\uparrow$ & LPIPS$_{v}$ $\downarrow$ \\
\hline
Mip-Splatting\textsubscript{4D}\cite{mip-splatting} & $37.76$ & $0.984$ & $0.020$ \\
Mip-Splatting\textsubscript{4D}\cite{mip-splatting} + M  &  38.20 & \second 0.985 &  0.018 \\ 
Mip-Splatting\textsubscript{4D}($\rho_{min} \cdot \sigma_s$)\cite{mip-splatting} + M  &  \third 38.32 & \second 0.985 & \best 0.017 \\

\hline
Ours w/o $\mathcal{L}_{\text{scales}}$ & \best 38.47 & \best 0.986 & \best 0.017 \\
Ours & \second 38.39 & \second 0.985 & \best 0.017 \\
\hline
\end{tabular}
}
\caption{\textbf{Ablation Study on the D-NeRF Dataset}~\cite{dnerf}. Results show that incorporating the mask $M$ to prevent the dilation of invisible Gaussians improves reconstruction quality. Furthermore, even when adjusting Mip-Splatting’s expansion scale to match the minimum dilation scale of our method, its performance remains inferior.}
\label{tab:abl}
\end{table}

\begin{figure}[tb] 
\centering
\includegraphics[width=1.0\columnwidth]{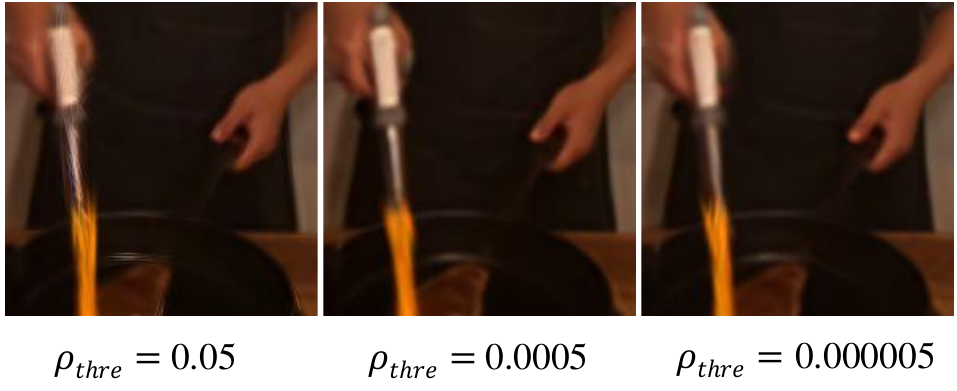} 
\caption{\textbf{Ablation Study on the Hyperparameter} $\rho_{\text{thre}}$: A lower $\rho_{\text{thre}}$ imposes stricter frequency constraints on the Gaussian while reducing sensitivity to illumination variations.}
\label{fig:thre_abl}
\end{figure}

\section{Experiments}
We first present the implementation details of our method. We then evaluate its performance on the monocular video reconstruction dataset, D-NeRF~\cite{dnerf}, and the multi-view video reconstruction dataset, Neural 3D Video Dataset (N3DV)~\cite{n3dv}. Finally, we discuss the limitations of our approach.

\subsection{Implementation}
We integrated our method into two state-of-the-art reconstruction frameworks: \textbf{D3DGS}~\cite{d3dgs}, specialized for monocular video reconstruction, and \textbf{4DGaussians}~\cite{4dGaussians}, a deformation-based multi-view approach. Both the 2D Mip filter~\cite{mip-splatting} and the 4D scale-adaptive filter were employed. Parameter settings followed their respective baselines, except for the densification threshold $\tau_g$ in \textbf{D3DGS}, which was adjusted to $0.0006$. To maintain consistency with Mip-Splatting~\cite{mip-splatting}, we set $\sigma_s = 0.2$. For a fair generalization assessment, we used uniform settings of $\rho_{\text{min}} = 0.2$ and $\rho_{\text{max}} = 5$ across both frameworks, ensuring that the minimum effective Gaussian range approximated one pixel (defined as three times the Gaussian scale). When the rendering sampling rate decreases, we increase $\rho_{\text{min}}$ by multiplying it by the square of the ratio between the current and initial sampling rates, strictly capping it at 1. This adjustment is crucial, as a lower $\rho_{\text{min}}$ reduces the dilation scale, leading to a loss of Gaussian blurring on high-frequency information at lower sampling rates. To counteract this effect, we apply corresponding adjustments during rendering. The threshold $\rho_{\text{thre}}$ was task-dependent, set to $0.05$ for D3DGS and $5 \times 10^{-6}$ for 4DGaussians. Additionally, we set $\lambda_v = 0.2$ and $\lambda_1 = 0.1$. Notably, in 4DGaussians, $\mathcal{L}_{\text{scales}}$ was computed using an average based on $M_1$.

% We integrated our method into two state-of-the-art approaches: \textbf{D3DGS} \cite{d3dgs}, which excels in monocular video reconstruction, and \textbf{4DGaussians} \cite{4dGaussians}, a deformation-based multi-view video reconstruction method. Both the 2D Mip filter \cite{mip-splatting} and the 4D scale-adaptive filter were employed. All parameters remained consistent with the respective baseline implementations, except for \textbf{D3DGS}, where we adjusted the densification threshold $\tau_g$ to 0.0006. In our experiments, we set $\sigma_s$ to $0.2$ to align with Mip-Splatting\cite{mip-splatting}. 为了验证我们方法的通用性，在D3DGS和4DGaussians The parameters $\rho_{\text{min}}, \rho_{\text{max}},$ 都were set to $0.2$, $5$.请注意，我们简单任务高斯的有效范围是高斯尺度的3倍，$0.2$是 4D scale-adaptive filter和scale Regularization联合约束下保证高斯最小有效范围逼近1个像素的最小值。
%  我们通过设置不同的$\rho_{\text{thre}}$ 适应不同的表征和任务, $\rho_{\text{thre}}$ 在D3DGS和4DGaussians中分别设置成 $0.05$、$0.000005$, respectively, while $\lambda_v$ and $\lambda_1$ were set to $0.2$ and $0.1$.不同的是在4DGaussians计算$\mathcal{L}_{\text{scales}}$时基于$M1$做了加权平均。

\subsection{Evaluation on D-NeRF dataset}
\textbf{Multi-scale Training and Testing:} Following previous works \cite{mip-nerf, tri-miprf, dmit}, we downsample images from the D-NeRF dataset by factors of 2, 4, and 8 to obtain multi-scale observations, adjusting focal lengths accordingly based on perspective projection. This setup facilitates a comprehensive anti-aliasing evaluation by computing metrics across scales, as summarized in \cref{tab:dnerf_multi_train}. Our method consistently outperforms all prior approaches across all metrics, including DMiT \cite{dmit}, which utilizes the Mipmapped Tri-Plane representation.

\noindent \textbf{Single-scale Training and Multi-scale Testing:} Following~\cite{mip-splatting, sa-gs}, we train on full-resolution images and render at multiple scales (\(1\times\), \(1/2\), \(1/4\), and \(1/8\)) to simulate zoom-out effects. Given the absence of publicly available benchmarks for 4D Gaussian Splatting in this context, we reproduce D3DGS~\cite{d3dgs} by replacing its 2D dilated filter with a 2D Mip filter~\cite{mip-splatting}. Additionally, we adapt Mip-Splatting to 4DGS by introducing our proposed 4D frequency computation method in place of the original. To assess the effectiveness of our approach, we compare it against these implementations. Notably, our \emph{4D Scale-adaptive Filter} can be configured to match the Mip-Splatting-based 4DGS implementation under specific parameters while offering greater flexibility.Quantitative comparisons are presented in~\cref{tab:dnerf_ds}. Although Mip-Splatting demonstrates good anti-aliasing capabilities, it reduces the reconstruction quality of D3DGS at full resolution. Qualitative comparisons in~\cref{fig:dnerf_ds1} show that D3DGS effectively captures deformation details at full resolution but introduces significant inflation artifacts when tested at lower sampling rates. After integrating Mip-Splatting, the capability of D3DGS to capture fine deformation details deteriorates. In contrast, our method maintains the original reconstruction quality of D3DGS and preserves more realistic scene details at lower resolutions.

% \cref{tab:dnerf_ds}展现了定量化结果，虽然Mip-Splatting展现了不错的抗锯齿能力，但会降低D3DGS在全分辨率下的重建质量。\cref{fig:dnerf_ds1}提供了定性比较，D3DGS能够在全分辨率下捕获变形细节，但是当降低采样率时表现出了明显的膨胀伪影。集成Mip-Splatting后，D3DGS在捕获变形细节的能力有一定的下降。我们的方法在不降低D3DGS重建质量的同时在低分辨率下保留了更真实的场景细节。

% Following \cite{mip-splatting, sa-gs},
% training on full-resolution images and rendering at various resolutions (i.e. 1×, 1/2, 1/4, and 1/8) to mimic zoom-out effects.由于这个设置在4DGS缺乏公开的基准，我们复现了D3DGS\cite{d3dgs}，只用2D Mip filter\cite{mip-splatting}替换掉D3DGS中的2d dilated filter，以及将Mip-Splatting迁移到D3DGS中实现了Mip-Splatting的4DGS版本，具体来说是用我们提出的4D中的最大频率计算方法替换掉了原来Mip-Splatting的最大频率计算方法。并让我们的方法与这些实现做对比来验证我们方法的有效性，请注意，我们提出的$4D Scale-adaptive Filter$设置特定参数可以和Mip-Splatting的4DGS版本完全相同，但$4D Scale-adaptive Filter$比之要更加灵活.

\subsection{Evaluation on N3DV dataset}
\noindent \textbf{Single-scale Training and Multi-scale Testing:} To simulate zoom-in effects, we define the default resolution of 4DGaussians (1352×1014) as the full resolution. Models are trained on data downsampled by a factor of 4 and rendered at progressively higher resolutions (1×, 2×, and 4×). Following the approach used in the D-NeRF dataset, we integrate Mip-Splatting~\cite{mip-splatting} into 4DGaussians as a baseline for comparison, ensuring consistency in the maximum sampling frequency computation between Mip-Splatting and our method. The results in \cref{tab:dynerf_up} demonstrate that our method achieves anti-aliasing performance comparable to $Mip$-$Splatting$ while effectively constraining the frequency of 4DGS, reducing redundant Gaussians, and improving reconstruction quality. As shown in \cref{fig:dynerf_up1}, our approach preserves the ability of 4DGS to learn deformations but affects its sensitivity to lighting. This effect arises from the scales loss constraining the intrinsic scale of Gaussians, which can be mitigated by adjusting $\rho_{\text{thre}}$, a more in-depth analysis is provided in the supplementary materials.

% \cref{tab:dynerf_up}中的结果显示我们的方法与 $Mip-Splatting$ 有相当的抗锯齿能力，并且对4DGS的频率进行有效约束后，能够减少冗余高斯，提升重建质量。如\cref{fig:dynerf_up1}所示，我们的方法不会降低4DGS学习变形的能力，但会影响对光照的敏感度，这是由于scales loss去约束高斯自身的尺度的原因，可以通过调节$\rho_{\text{thre}}$来缓解. 

% To simulate zoom-in effects, 我们将4DGaussians的默认分辨率 1352×1014作为全分辨率，we train models on data downsampled by a factor of 4 and rendered at successively higher resolutions (1×, 2×, and 4x).与在D-NeRF dataset中一样，我们将Mip-Splatting\cite{mip-splatting}集成到4DGaussians中作为对比方法，Mip-Splatting中的最大采样频率计算方法与我们保持一致。

\subsection{Ablation Study}
\noindent \textbf{4D Scale-adaptive Filter:} Compared with the 3D smoothing filter~\cite{mip-splatting}, the proposed 4D scale-adaptive filter has two advantages in 4DGS: (1) it prevents unnecessary splatting of invisible Gaussians, and (2) it preserves Gaussian anisotropy by adaptively adjusting splatting scales within a defined range. To validate these advantages, we integrate the masking operation $M$ from \cref{eq:mask} into Mip-Splatting to avoid splatting invisible Gaussians and set its fixed scale equal to the minimal adaptive scale $\rho_{min} \cdot \sigma_s$ used by the proposed method. Quantitative results in \cref{tab:abl} demonstrate the effectiveness of the 4D scale-adaptive filter beyond merely using smaller splatting scales.

% 4D Scale-adaptive Filter对比3D smoothing filter\cite{mip-splatting}在4DGS中的优势是两点，一是避免对不可见高斯的膨胀。二是在通过在一定范围内调整膨胀尺度来减少滤波操作对高斯各向异性的影响。为了验证这两点的有效性，我们在Mip-Splatting上同样加入\cref{eq:mask}中的判定操作$M$来避免对不可见的高斯的膨胀，此后在调整其固定膨胀尺度为4D Scale-adaptive Filter的最小膨胀尺度$\rho_{min} \cdot \sigma_s$,以此来证明4D Scale-adaptive Filter的有效不只是来自于一个更小的膨胀尺度，\cref{tab:abl}中的定量结果证明了4D Scale-adaptive Filter的优势。

\noindent \textbf{Scale Regularization:}To validate the effectiveness of jointly employing the scale loss $\mathcal{L}_{\text{scales}}$ and the 4D scale-adaptive filter for constraining frequencies of 4D Gaussians, we train models on the D-NeRF dataset~\cite{dnerf}, downsampled by a factor of 4, and evaluate rendering quality at progressively higher resolutions (1$\times$, 2$\times$, and 4$\times$). Quantitative results are summarized in \cref{tab:dner_up}. Due to the negative impact of the D3DGS filter~\cite{d3dgs} on full-resolution reconstruction, we adopt a smaller minimum dilation ratio $\rho_{\text{min}}$, which, however, yields insufficient frequency constraints and poorer anti-aliasing compared to Mip-Splatting~\cite{mip-splatting} at higher resolutions. By introducing $\mathcal{L}_{\text{scales}}$ to regularize Gaussian scales during deformation, our approach effectively alleviates this limitation, achieving superior anti-aliasing performance over Mip-Splatting in monocular video reconstruction tasks.

% 为了验证scales loss $\mathcal{L}_{\text{scales}}$联合4D scale-adaptive filter对4DGS的频率的有效约束，我们train models on D-NeRF dataset\cite{dnerf} downsampled by a factor of 4 and rendered at successively higher resolutions (1×, 2× and 4×)，由于为了减少在D3DGS\cite{d3dgs}滤波器对全分辨率下重建的影响，我们将4D scale-adaptive filter最小膨胀比例$\rho_\text{min}$设置的比较小，对高斯频率约束不足，在高分辨率下的效果不如Mip-splatting\cite{mip-splatting}，我们通过用$\mathcal{L}_{\text{scales}}$来Regularization高斯变形过程中的scale，在单目视频重建任务中抗锯齿能力超过了Mip-Splatting.

%% file: sec/6_conclusion.tex
\section{Limitations}
Our method is sensitive to the hyperparameter $\rho_{\text{thre}}$, which determines the critical scale at which Gaussians are prevented from dilation and subjected to scales loss constraints. As shown in \cref{fig:thre_abl}, a smaller $\rho_{\text{thre}}$ reduces 4DGS's sensitivity to lighting, while a larger $\rho_{\text{thre}}$ weakens frequency constraints on 4DGS. Additionally, if a Gaussian fails to deform accurately, scale regularization may enlarge redundant small Gaussians, leading to negative optimization.

\section{Conclusion}
We propose a maximum sampling frequency formulation for 4DGS and introduce the 4D Scale-Adaptive Filter and Scale Regularization to effectively constrain Gaussian frequencies. Experiments demonstrate that in monocular video reconstruction, our method eliminates severe artifacts caused by sampling rate changes without compromising reconstruction quality. In multi-view video reconstruction, it not only removes these artifacts but also significantly reduces redundant Gaussians, improving reconstruction quality.

% 我们提出了4DGS的最大采样频率计算公式，并基于它提出 4D Scale-adaptive Filter 和 Scale Regularization来对高斯的频率进行有效约束。实验证明，我们的方法在单目视频重建任务中，在不影响重建效果的前提下，消除采样率改变时产生的强烈伪影。而在多目视频重建任务中，我们的方法不仅能消除采样率改变时产生的强烈伪影，而且能够有效减少冗余高斯，提高重建质量。

%% file: sec/X_suppl.tex
\clearpage
\setcounter{page}{1}
\maketitlesupplementary
\begin{abstract}
  \noindent
  This supplementary material accompanies the main paper by providing more details
  for reproducibility as well as additional evaluations and qualitative results to
  verify the effectiveness and robustness of \ours:\\
  \noindent
  $\triangleright$ \textbf{\cref{sec:impl_detail}}: Additional implementation details.
  \\
  \noindent
  $\triangleright$ \textbf{\cref{sec:add_exp}}: Additional experimental results,
  including more detailed view synthesis quality comparison, rendering visualization at multiple scales.
\end{abstract}

% \section{Rationale}
% \label{sec:rationale}
% %
% Having the supplementary compiled together with the main paper means that:
% %
% \begin{itemize}
% \item The supplementary can back-reference sections of the main paper, for example, we can refer to \cref{sec:intro};
% \item The main paper can forward reference sub-sections within the supplementary explicitly (e.g. referring to a particular experiment);
% \item When submitted to arXiv, the supplementary will already included at the end of the paper.
% \end{itemize}
% %
% To split the supplementary pages from the main paper, you can use \href{https://support.apple.com/en-ca/guide/preview/prvw11793/mac#:~:text=Delete%20a%20page%20from%20a,or%20choose%20Edit%20%3E%20Delete).}{Preview (on macOS)}, \href{https://www.adobe.com/acrobat/how-to/delete-pages-from-pdf.html#:~:text=Choose%20%E2%80%9CTools%E2%80%9D%20%3E%20%E2%80%9COrganize,or%20pages%20from%20the%20file.}{Adobe Acrobat} (on all OSs), as well as \href{https://superuser.com/questions/517986/is-it-possible-to-delete-some-pages-of-a-pdf-document}{command line tools}.
\section{Implementation Details}
\label{sec:impl_detail} \textbf{For monocular video reconstruction}, we adopt D3DGS~\cite{d3dgs},
the state-of-the-art method, as our codebase. Similar to Mip-Splatting~\cite{mip-splatting},
we implement a coarse minimum sampling interval estimation~\cref{eq:static_max_t}
and a 4D scale-adaptive filter in PyTorch, while a more accurate minimum sampling
interval update~\cref{eq:min_T} and a 2D Mip Filter~\cite{mip-splatting} are implemented
in CUDA. During the initial 3k iterations, we train only the 3D Gaussians to achieve
stable positions and shapes. Subsequently, joint optimization of 3D Gaussians and
the deformation field is performed. The minimum sampling interval $\hat{T}_{k}$
is computed using~\cref{eq:static_max_t} for the first 6k iterations and then
refined with~\cref{eq:min_T}. The maximum sampling frequency is the inverse of~\cref{eq:min_T},
ensuring consistency with D3DGS. The total training consists of 40k iterations.

\textbf{For multi-view video reconstruction}, we adopt 4DGaussian~\cite{4dGaussians}, another
deformation-based approach, as our codebase, with minimal modifications except for
the incorporation of an averaging operation using $M1$ in the scale loss~\cref{eq:scales_loss}.
Following 4DGaussian, we train only the 3D Gaussians for the first 3k iterations
to stabilize their positions and shapes before optimizing the deformation field.
Given the severe overfitting of 4DGaussian on the N3DV dataset~\cite{n3dv}, we limit
the deformation field training to 20k iterations, evaluating every 1k iterations
to select the best checkpoint for final assessment. Notably, all models in our
experiments follow this protocol to ensure fair comparisons.

\section{Additional Experimental Results}
\label{sec:add_exp}

\begin{figure*}[bt]
  \centering
  \includegraphics[width=1.0\textwidth]{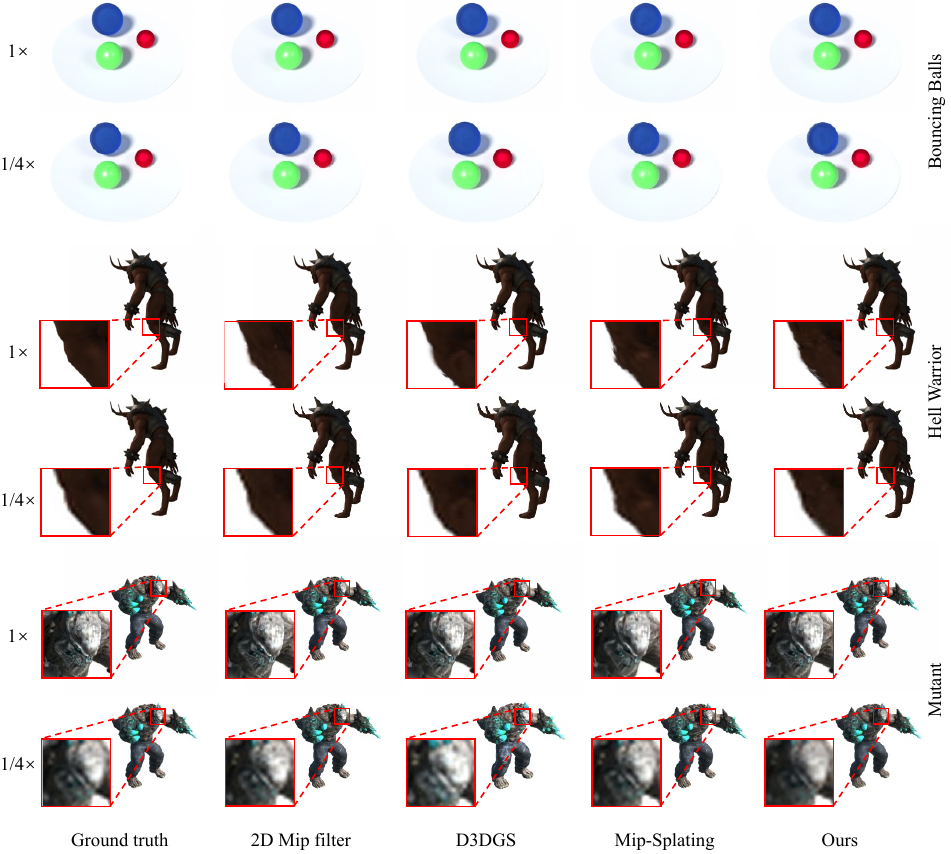}
  \caption{Single-scale Training and Multi-scale Synthesis on the D-NeRF dataset~\cite{dnerf}.All
  methods are trained at full resolution and evaluated at various lower
  resolutions to simulate zoom-out effects. D3DGS~\cite{d3dgs} exhibits noticeable
  blurring and inflation artifacts at lower resolutions. Integrating Mip-splatting~\cite{mip-splatting}
  mitigates inflation but introduces local reconstruction distortions. In contrast,
  our method preserves the reconstruction quality of D3DGS while maintaining
  more realistic visual fidelity at lower resolutions.}
  \label{fig:dnerf_ds1_supp}
\end{figure*}

\boldparagraph{Single-scale Training and Multi-scale Synthesis on the D-NeRF~\cite{dnerf} dataset.}
As shown in~\cref{fig:dnerf_ds1_supp}, we train all methods on full-resolution images and evaluate them at scale \(1\times\) and scale \(1/4\) to simulate zoom-out effects. Both 4DGS with Mip-filter and Mip-Splatting~\cite{mip-splatting} exhibit noticeable blurring and inflation artifacts at lower resolutions, especially in the Hell Warrior scene. While D3DGS~\cite{d3dgs} captures fine deformation details at full resolution, it introduces significant inflation artifacts at lower resolutions. In contrast, our method maintains the original reconstruction quality of D3DGS and preserves more realistic scene details at lower resolutions.

\boldparagraph{Single-Scale Training and Multi-Scale Evaluation on the N3DV Dataset~\cite{n3dv}.}
In the main text, we conducted simulated upscaling experiments on the N3DV dataset~\cite{n3dv}. Here, we further perform simulated downscaling experiments. \Cref{tab:dynerf_ds_supp} presents the quantitative results, showing that constraining Gaussian frequency reduces redundant Gaussians and slightly improves reconstruction quality. While our method demonstrates comparable anti-aliasing capabilities to Mip-Splatting~\cite{mip-splatting}, its full-resolution reconstruction quality is slightly lower than expected.

We analyze potential reasons for this discrepancy. In multi-view video reconstruction, the presence of redundant Gaussians, inaccurate motion modeling, and limited adaptability of Gaussian deformation may contribute to this outcome. Our proposed scales loss is designed to constrain visible Gaussians; however, if a Gaussian fails to deform correctly and become invisible, scales loss may act as a negative optimization. In contrast, Mip-Splatting applies large-scale dilation filtering, which significantly reduces Gaussian opacity after rendering and is less affected. Since scales loss does not regulate transparency, it may lead to small redundant Gaussians expanding instead of disappearing, resulting in adverse optimization effects.

\begin{figure*}[bt]
  \centering
  \includegraphics[width=1.0\textwidth]{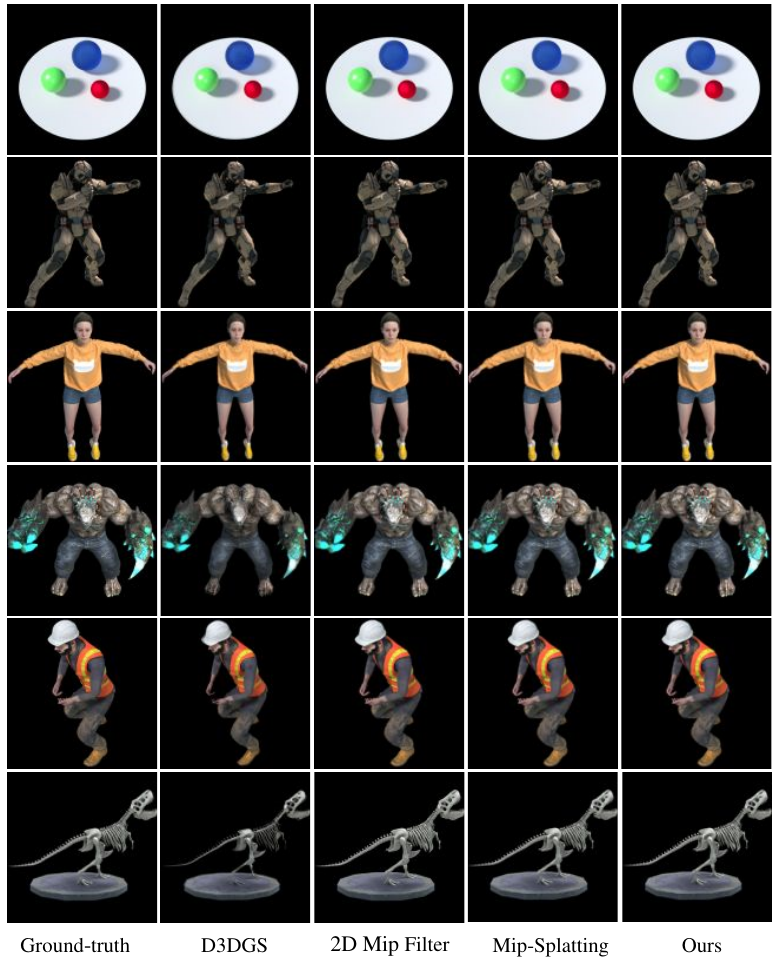}
  \caption{\textbf{Single-Scale Training and Multi-Scale Synthesis on the D-NeRF dataset~\cite{dnerf}}. All models are trained on images downsampled by a factor of four and rendered at full resolution to simulate zoom-in. Our method effectively eliminates high-frequency artifacts and produces more complete object shapes compared to Mip-Splatting.}
  \label{fig:dnerf_black_supp}
\end{figure*}

\boldparagraph{Single-Scale Training and Multi-Scale Rendering on the D-NeRF Dataset~\cite{dnerf}.}
We conduct additional experiments on the D-NeRF dataset~\cite{n3dv} to evaluate the performance of our method at different scales. As shown in~\cref{fig:dnerf_black_supp}, we train all models on images downsampled by a factor of four and render them at full resolution to simulate zoom-in and moving-closer effects. Our method effectively eliminates high-frequency artifacts and produces more complete object shapes compared to other methods. In contrast, D3DGS~\cite{d3dgs} exhibits noticeable blurring and inflation artifacts. D3DGS~\cite{d3dgs} integrating Mip-Splatting~\cite{mip-splatting} and the 2D Mip filter achieves better visual fidelity but exhibits numerous high-frequency artifacts. In additional, Mip-Splatting~\cite{mip-splatting} fails to capture fine details in dynamic reconstruction and produces incomplete object shapes.

\begin{table*}
  [t]
  \centering
  \resizebox{\textwidth}{!}{%
  \begin{tabular}{l|ccccc|ccccc|ccccc|c}
    \hline
    \multirow{2}{*}{\textbf{Methods}}                    & \multicolumn{5}{c|}{\textbf{PSNR $\uparrow$}} & \multicolumn{5}{c|}{\textbf{SSIM $\uparrow$ }} & \multicolumn{5}{c|}{\textbf{LPIPS$_{v}$ $\downarrow$}} & \multirow{2}{*}{\textbf{$|\mathcal{G}|$}} \\
    \cline{2-6} \cline{7-11} \cline{12-16}               & Full Res.                                     & 1/2 Res.                                       & 1/4 Res.                                               & 1/8 Res.                                 & Avg.          & Full Res.     & 1/2 Res.      & 1/4 Res.      & 1/8 Res.    & Avg.          & Full Res.     & 1/2 Res.      & 1/4 Res.      & 1/8 Res.    & Avg.          &             \\
    \hline
    4DGaussian \cite{4dGaussians} *                      & 31.56                                         & 29.49                                          & 25.99                                                  & 22.46                                    & 27.38         & \best 0.938   & 0.916         & 0.858         & 0.763       & 0.869         & \best 0.148   & 0.106         & 0.094         & 0.124       & 0.118         & 125k        \\
    2D Mip Filter \cite{mip-splatting}                   & 31.66                                         & 32.06                                          & 31.90                                                  & 30.50                                    & 31.53         & \second 0.936 & \second 0.945 & 0.955         & 0.950       & 0.946         & \second 0.150 & \best 0.096   & \best 0.045   & 0.032       & \best 0.081   & 103k        \\
    Mip-Splatting\textsubscript{4D} \cite{mip-splatting} & \best 31.86                                   & \best 32.35                                    & \best 32.57                                            & \best 31.54                              & \best 32.08   & 0.937         & \second 0.945 & \second 0.956 & \best 0.955 & \second 0.948 & 0.155         & \second 0.099 & \second 0.047 & \best 0.031 & \second 0.083 & \second 94k \\
    \hline
    Ours                                                 & \second 31.75                                 & \second 32.22                                  & \second 32.35                                          & \second 31.28                            & \second 31.90 & 0.937         & \best 0.946   & \best 0.957   & \best 0.955 & \best 0.949   & 0.154         & \second 0.099 & \second 0.047 & \best 0.031 & \second 0.083 & \best 92k   \\
    \hline
  \end{tabular} %
  }
  \caption{\textbf{Single-Scale Training and Multi-Scale Evaluation on the N3DV Dataset}~\cite{n3dv}.}
  \label{tab:dynerf_ds_supp}
\end{table*}

